\documentclass[graybox]{svmult}

\usepackage{mathptmx}       
\usepackage{helvet}         
\usepackage{courier}        
\usepackage{type1cm}        
\usepackage{makeidx}         
\usepackage{graphicx}        
\usepackage{multicol}        
\usepackage[bottom]{footmisc}

\makeindex             

\begin{document}

\title*{Evaluation of a semi-autonomous attentive listening system with takeover prompting}
\author{Haruki Kawai, Divesh Lala, Koji Inoue, Keiko Ochi and Tatsuya Kawahara}
\institute{Haruki Kawai \at Kyoto University Graduate School of Informatics, \email{kawai@sap.ist.i.kyoto-u.ac.jp}
\and Divesh Lala \at Kyoto University Graduate School of Informatics, \email{lala@sap.ist.i.kyoto-u.ac.jp}
\and Koji Inoue \at Kyoto University Graduate School of Informatics, \email{inoue@sap.ist.i.kyoto-u.ac.jp}
\and Keiko Ochi \at Kyoto University Graduate School of Informatics, \email{ochi.keiko.5f@kyoto-u.ac.jp }
\and Tatsuya Kawahara \at Kyoto University Graduate School of Informatics, \email{kawahara@i.kyoto-u.ac.jp}}
%
%
\maketitle

\abstract*{The handling of communication breakdowns and loss of engagement is an important aspect of spoken dialogue systems, particularly for chatting systems such as attentive listening, where the user is mostly speaking. We presume that a human is best equipped to handle this task and rescue the flow of conversation. To this end, we propose a semi-autonomous system, where a remote operator can take control of an autonomous attentive listening system in real-time. In order to make human intervention easy and consistent, we introduce automatic detection of low interest and engagement to provide explicit takeover prompts to the remote operator. We implement this semi-autonomous system which detects takeover points for the operator and compare it to fully tele-operated and fully autonomous attentive listening systems. We find that the semi-autonomous system is generally perceived more positively than the autonomous system. The results suggest that identifying points of conversation when the user starts to lose interest may help us improve a fully autonomous dialogue system.}

\abstract{The handling of communication breakdowns and loss of engagement is an important aspect of spoken dialogue systems, particularly for chatting systems such as attentive listening, where the user is mostly speaking. We presume that a human is best equipped to handle this task and rescue the flow of conversation. To this end, we propose a semi-autonomous system, where a remote operator can take control of an autonomous attentive listening system in real-time. In order to make human intervention easy and consistent, we introduce automatic detection of low interest and engagement to provide explicit takeover prompts to the remote operator. We implement this semi-autonomous system which detects takeover points for the operator and compare it to fully tele-operated and fully autonomous attentive listening systems. We find that the semi-autonomous system is generally perceived more positively than the autonomous system. The results suggest that identifying points of conversation when the user starts to lose interest may help us improve a fully autonomous dialogue system.}

\section{Introduction}
The goal of studies on spoken dialogue systems is to make unconstrained talk with conversational agents seem as human-like as possible. The agent should exhibit understanding of the user's talk by providing appropriate responses at the right moments. Furthermore, an ideal agent would be able to keep the attention of the user for a sustained period of time.

One advantage humans possess in regards to this aspect of conversation is the ability to quickly estimate the state of the conversation in terms of engagement or interest and modify the talk to steer it in a more satisfactory direction for both parties (assuming that is a shared goal). For agents, to identify exactly when a conversation is breaking down is still a challenge, let alone how to naturally manage this breakdown as part of a dialogue policy. However, it will be possible to handle these situations and recover the dialogue through a human who can manipulate the talk.

\begin{figure}[t]
\centering
\includegraphics[width=0.80\columnwidth]{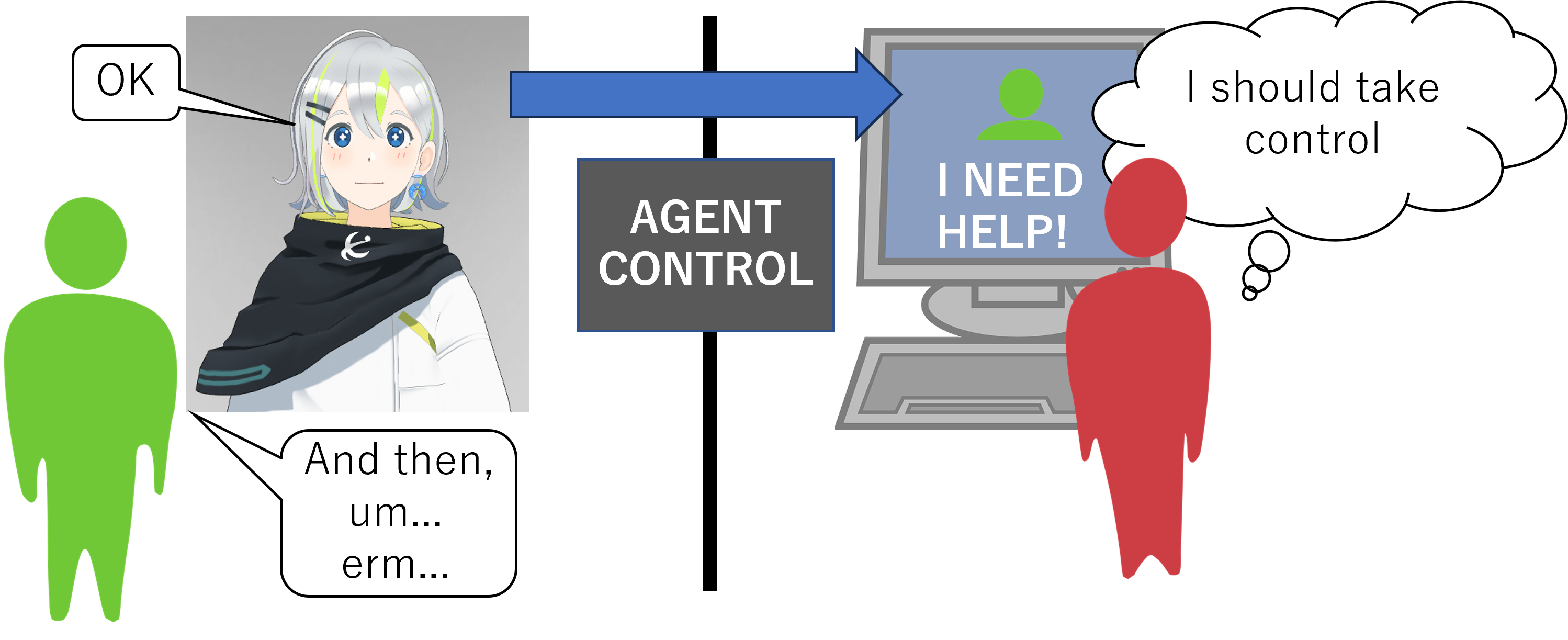}
\includegraphics[width=0.80\columnwidth]{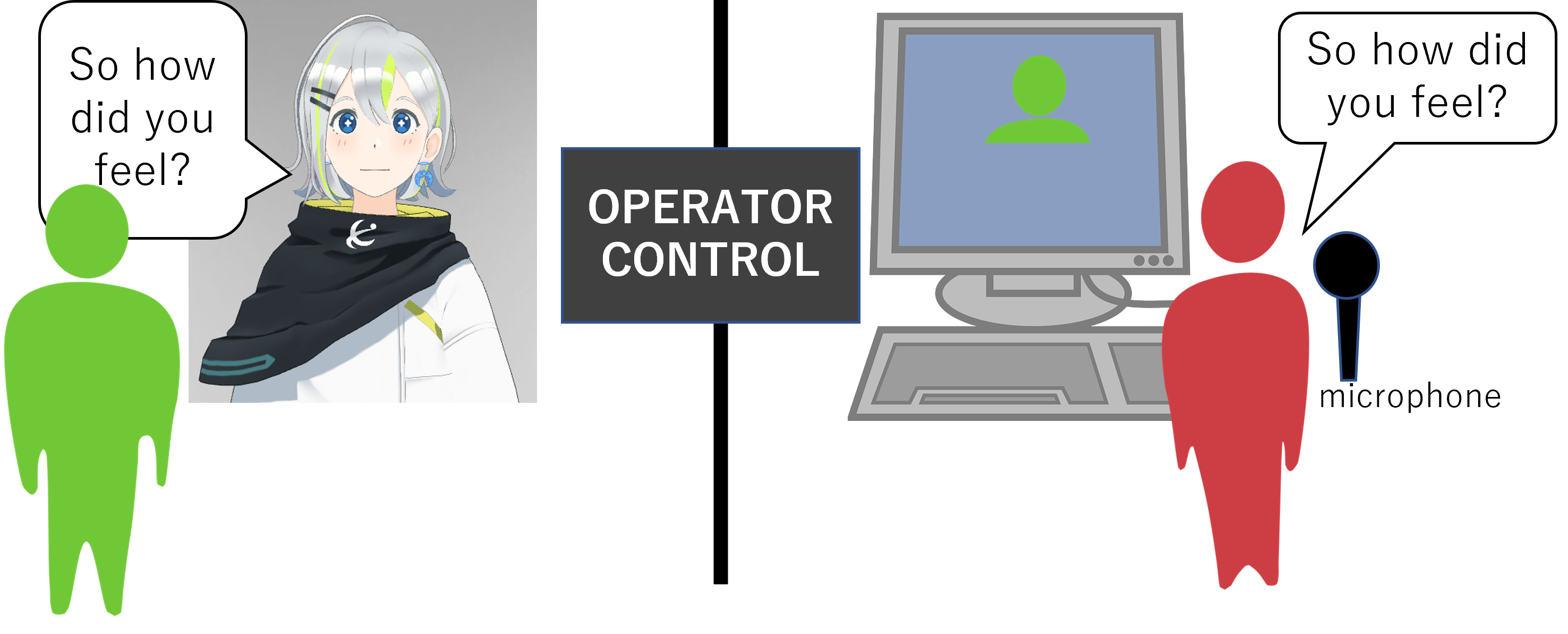}
\caption{General concept of the semi-autonomous system. The top half shows the dialogue system is in agent control, where it is autonomous and being monitored by a remote operator on the right hand side. If the system detects that there is a problem with the conversation then it informs the operator through the GUI who switches to operator control. The operator can then take over and speak through the agent. Switching between agent and operator control is done by the operator.}
\label{concept} 
\end{figure}

Towards this goal we propose a semi-autonomous system, where the generation of dialogue is mostly performed by the autonomous dialogue manager but a remote human operator is able to take over the conversation and speak through the agent. In a semi-autonomous system, the ``difficult'' parts of conversation can be handed off to a human expert while the autonomous system only needs to deal with situations which fit its purpose. The general concept of the system is shown in Fig. \ref{concept}. 

These types of handover systems are not new, and have been implemented in various scenarios \cite{Poser2021, Poser2022, Walker2000}. However, up until now these have mostly been restricted to service-type systems (e.g. call centers), where the dialogue system cannot handle a request from the user because it is out of domain. On the other hand, we are interested in a conversational scenario, where the decision to take over is not based on the domain knowledge of the agent but on the engagement level of the user. For a service-type system, the answer that the operator provides is related to the user's request. For a conversational system, the operator's ``answer'' could be more complex and is a means to keep the conversation flowing. 

The use of a semi-autonomous system also helps to achieve the goal of creating a better autonomous system. If we can identify the points of the conversation which are likely to benefit from human intervention, we can focus attention on implementing models which address these issues. For example, it may make sense to carefully use more powerful large language models (LLMs) in these situations. Through this method we can hopefully improve the overall system until we reach a point where the autonomous system is close to human-human conversation. 

This paper addresses the above issues by describing an experiment in which a remote human operator can intervene during the dialogue to improve the conversation. We identify specific points in the conversation where the autonomous system needs assistance with attentive listening. The system itself identifies these points and informs a remote operator through a special interface who can then take over on behalf of the autonomous system. We then compare this semi-autonomous system to two attentive listening systems which act as lower and upper baselines - a completely autonomous system and a completely tele-operated system. The target dialogue system in this work is in the Japanese language.

\section{Related Work}\label{related}
Semi-autonomous systems where a remote operator takes control of an autonomous system are not a new concept, and have been 
 implemented in service-oriented systems \cite{Poser2021,Poser2022} and spoken dialogue systems \cite{Walker2000}. However, the majority of these focus on task-based settings whereas we implement our system in a chat-style scenario. Recent works have analyzed the use of utterances after handover in a spoken dialogue system, but the scenario involved two human operators rather than a semi-autonomous system \cite{Yamashita2022}. One work \cite{Benner2021} states that handover to humans in spoken dialogue is rare even for dyadic interaction, so this presents research opportunities in this direction.

The issue of when the operator should take over is also one of our research goals, and we can see this as the task of identifying user engagement, which is a well-researched topic in dialogue systems \cite{Oertel2020}. Engagement has been defined in several ways including attentiveness towards a conversation \cite{Yu2004}, establishing and maintaining an interaction \cite{Sidner2005}, and how willing participants are to continue the dialogue \cite{Inoue2019}. Robust real-time engagement prediction is still a relatively difficult task and can depend on the setting. Although other works have engagement annotated by third-party passive observers \cite{Inoue2016,Oertel2015}, agreement across annotators is still somewhat inconsistent. Furthermore, engagement is often detected while a user is in a listening role. For attentive listening, the user is mostly speaking, so we must be able to estimate their interest level while they are in this state.

\section{Attentive Listening System}\label{att_list}
In this work we use our attentive listening dialogue system as the target scenario. The purpose of this system is to elicit talk from the user and make the agent show empathy and understanding towards the contents of the user's conversation. The user can speak with the system about any topic and the system will provide short responses. We define several types of system responses:

\begin{itemize}
\item{\textbf{Assessments}} reflecting some emotion (``That's great!'', ``That's a shame...')
\item{\textbf{Elaborating questions}} which use a keyword to generate a question for the user (``What type of ramen did you eat?'')
\item{\textbf{Repeated responses}} which repeat a keyword to show interest (``A train...'')
\item{\textbf{Formulaic responses}} which provide generic feedback (``I see..'', ``OK''). 
\end{itemize}

The choice of response to a user's turn is determined in a hierarchical manner according to the list above. For example, if both an elaborating question and a repeated response can be generated, then the former will take precedence and be uttered by the agent. The aim is to have the system often generate empathetic statements and follow-up questions as much as possible. The details of how these responses are generated and the attentive listening system itself are provided in previous work \cite{Lala2017}. The underlying concept of the system relies on detecting a focus word based on the user's dialogue and using this to generate elaborating questions and repeated responses. To prevent too many incorrect responses, the system only uses focus words for which the automatic speech recognition (ASR) model is sufficiently confident.

Additionally, the system also provides listening behavior in the form of backchannels, generated by a logistic regression model \cite{Lala2017}. The system can produce formal backchannels (such as \textit{un}, \textit{unun} and \textit{ununun}) as well as reactive backchannels (such as \textit{ah}, \textit{he-} and \textit{oh-}) which contain some emotional meaning. Reactive backchannels are produced when the user has said something with a positive or negative sentiment. If the user is silent for more than 5 seconds, the system will prompt them to continue talking with exploratory questions. 

\begin{figure}[t]
\centering
\includegraphics[width=0.50\columnwidth]{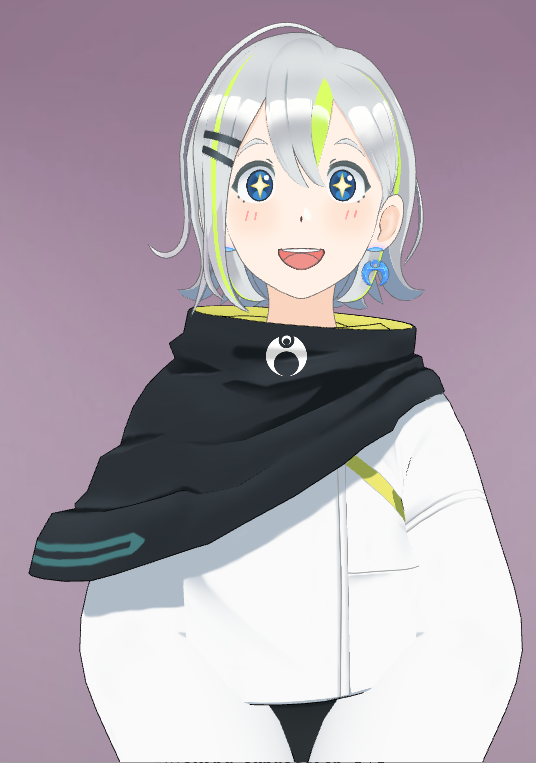}
\caption{The virtual agent used in the work, executing a ``happy'' expression.}
\label{agent} 
\end{figure}

The interface of the system is the virtual agent MMDAgent \cite{Lee2013} shown in Fig. \ref{agent}. The agent has automatic idling behavior, uses a special text-to-speech system (TTS), and is controlled by the above dialogue manager. For utterances which contain sentiments, the agent's facial expression becomes animated to display that particular emotion.

Users communicate with the agent through a stand microphone. ASR for Japanese is developed by our research group for this task \cite{Ueno2018}. The system also has voice activity detection and a turn-taking algorithm \cite{Lala2018} to control the timing of the agent responses.

An example of attentive listening dialogue with a user is as follows (translated into English):

\begin{itemize}
\item[]{\textbf{User}} I went for a really fun week-long trip to the Philippines
\item[]{\textbf{System}} \textit{Ah that's great!} (assessment)
\item[]{\textbf{User}} We spent our time looking at beaches and islands 
\item[]{\textbf{System}} \textit{yeah }(backchannel)
\item[]{\textbf{User}} and we also explored these places by bike
\item[]{\textbf{System}} \textit{By bike }(repeated response)
\end{itemize}

We see that most of the talk is done by the user, with the system providing short responses to provide conversational support.

\subsection{Semi-autonomous attentive listening}
The attentive listening system can be extended to a semi-autonomous attentive listening system, which allows a remote operator to take over and control MMDAgent's speech and facial expressions. It should be noted that the dialogue system itself is exactly the same as the autonomous system. The only difference is that the semi-autonomous system allows the remote operator to interrupt the conversation and speak to the user directly, and then relinquish control back to the autonomous system.

A video feed of the user is displayed to the operator who can also directly listen in on the conversation. Dialogue history is provided to the operator, with user dialogue being the result of ASR. The amount of user silence is also shown in the form of a progress bar. The user is able to take control of the agent by clicking a microphone icon. Once they have taken over the agent, anything that is said by the operator will be transmitted to the user who will hear their voice which has been converted to try and closely match the TTS voice. Lip synchronization is also implemented to match the operator's speech. To relinquish control of the agent back to the autonomous system, the operator simply clicks the microphone button again to toggle it off.

The operator can also manipulate the facial expression of the agent by clicking one of the provided emoji buttons (happy, sad, anger, surprise or laughter). Utterances can be output synchrounously with facial expressions. This allows the operator to communicate in a multi-modal manner. 

\begin{figure}[t]
\centering
\includegraphics[width=0.95\columnwidth]{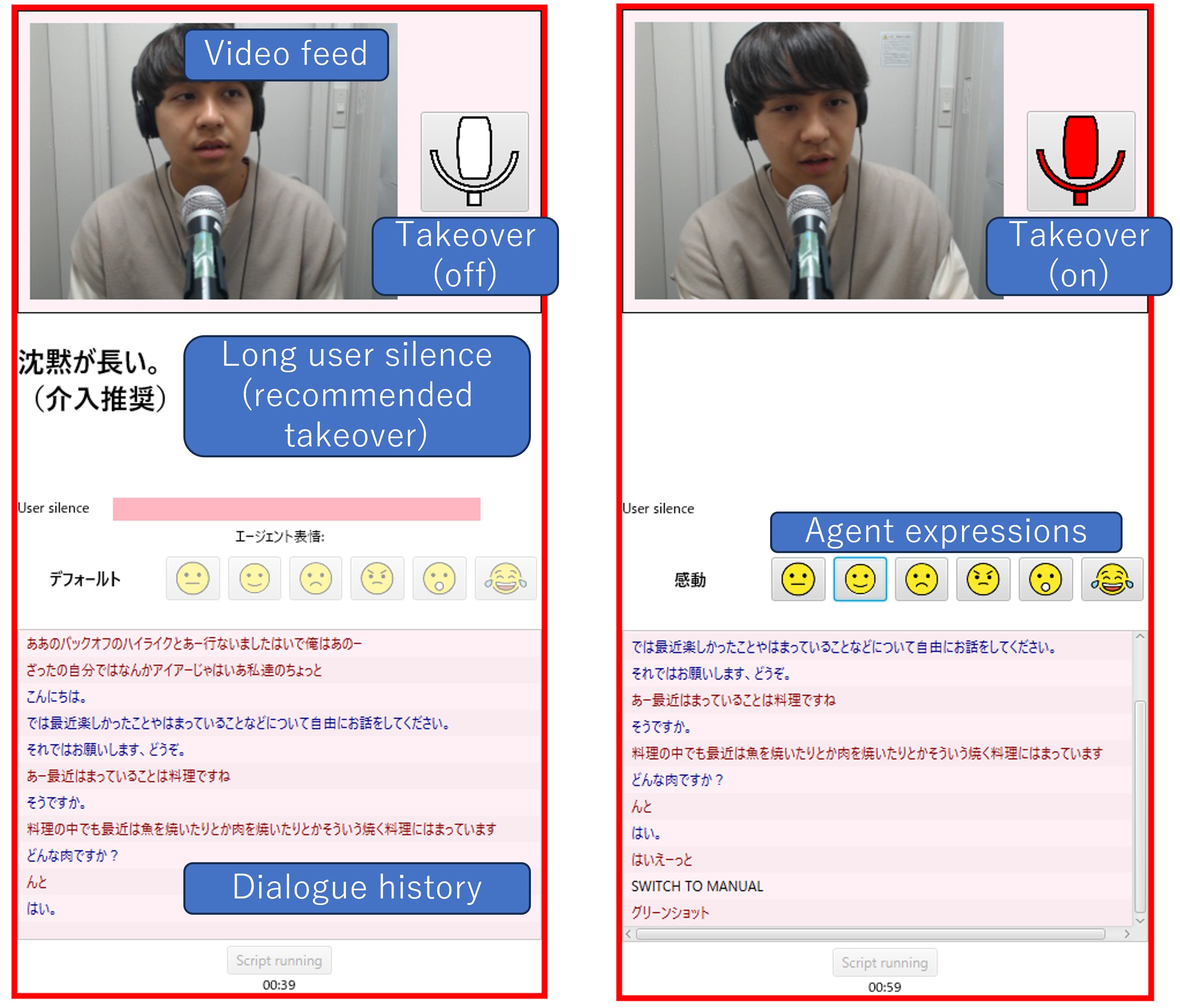}

\caption{The interface seen by the remote operator before takeover (left) and after takeover (right). Note that a message appears when the system detects that a takeover is necessary. Once the operator takes over, the microphone button is highlighted to show that they can directly speak to the user.}
\label{op_screen} 
\end{figure}

The interface change when taking over is shown in Fig. \ref{op_screen}. The left screenshot shows a user who has become silent for a long period of time, indicated by the user silence bar being colored red. The interface prompts the operator to take over and explains the reason why they should do so. Once the operator clicks the microphone button, the interface becomes the right screenshot. The emoji buttons are now activated and the microphone is highlighted, indicating that the operator can directly speak to the user.

\subsection{Preliminary experiments}
Initially we designed the system so that the remote operator can be free to choose when to take control, expecting that a human would best understand when communication was breaking down and steer the flow of the conversation. In the initial version of the interface, there was no prompt displayed which encouraged the operator to take over. 

Using this system we conducted several pilot experiments where operators listened to the attentive listening dialogue and were instructed to take over when they personally felt the user was not engaged or the conversation was breaking down. However, it was found that operators were reluctant to take over, with nearly half not even interjecting once. This was the case even when it seemed obvious that the conversation had broken down, such as lengthy periods of silence. Furthermore, the results showed that this type of system fared no better than the fully autonomous system in terms of user perception.

We hypothesized several reasons for the lack of activity by the operators. Firstly, the situation itself was a new one and the operators were unable to quickly familiarize themselves with the system. Secondly, social norms may have prevented operators from interrupting a conversation. Finally, and contrary to our initial assumption, some operators may simply be unable to use their own judgment to identify when to take over.

\subsection{Takeover prompting}
From the results of these initial experiments, it was clear that leaving the decision to take over completely to the operator would not be the optimal strategy for a semi-autonomous system. We introduced a mechanism to trigger the operator to take over, with visually explicit prompts. The system had to detect some signals which indicated that the conversation was breaking down or not engaging for the user.

Although several studies described in Section \ref{related} have been done for detection of engagement based on machine learning, many were inapplicable due to the recognition of user engagement as a listener. We adopt an explainable model so that the operator could know exactly why the system had made the choice to yield to them. The following represent conditions which indicate a potential lack of engagement or interest:

\begin{itemize}
\item Silence from the user for more than four seconds
\item Two consecutive turns from the user are less than 20 characters (phonemes) long
\item Three consecutive agent responses are formulaic
\item The last four agent responses have not contained any sentiment or elaborating questions
\end{itemize}

Silence was used as a condition since it is a simple proxy for engagement. If the user is silent we assume they cannot think of a suitable response or do not want to continue the line of conversation. 

Given that we want to encourage users to talk more, we also decided that the operator should intervene if the user's turn is too short. Our reasoning is that the user's turn is short because the questions or responses uttered by the agent were not interesting enough to convince the user to talk more. An example of this dialogue is given below:

\begin{itemize}
\item[]{\textbf{System}} \textit{What type of food did you eat?}
\item[]{\textbf{User}} Pasta carbonara.
\item[]{\textbf{System}} \textit{Carbonara?}
\item[]{\textbf{User}} It was fine.
\item[]{\textbf{System}} \textit{OK (sends a short turn notification to the operator)}
\end{itemize}

The final two conditions were chosen to try and maintain diversity within the agent responses, as we felt if the response patterns became repetitive then the user would lose interest. From previous work we know that sentiment responses are associated with pleasure \cite{Ochi2023}. Elaborating questions should also be encouraged since they require users to think and respond. This condition prioritizes both these response types. An example of such a dialogue is below:

\begin{itemize}
\item[]{\textbf{User}} So we went to the store and did some window shopping for a bit
\item[]{\textbf{System}} \textit{OK.}
\item[]{\textbf{User}} Then we chatted for a bit but it was closing time so we left
\item[]{\textbf{System}} \textit{I see.}
\item[]{\textbf{User}} It got pretty cold in the evening
\item[]{\textbf{System}} \textit{Yes (sends a notification about formulaic responses to the operator)}
\end{itemize}

This situation generally occurs when the attentive listening system cannot confidently generate a suitable elaborating question. This is due to either the output of the ASR being incorrect or focus words in the user's dialogue not meeting the threshold of ASR confidence.

It is possible for multiple conditions to be met at the same time (for example, short turns where the system uses too many formulaic responses). In this case, the system will just display two of the detected conditions to the operator. 

The specific values used for these conditions were formulated by analyzing previous sessions of attentive listening. We applied the values of these conditions to previously recorded interaction sessions of users with the autonomous system and found them to be present in most sessions but at a rate where the majority of the talk is still done by the autonomous system. However, it is impossible to determine the course of the conversation if a takeover were to occur, and if takeover would improve the system. This motivates the need for the experiments described in this paper. 

\section{Experiments}
We performed three experiments in this work to evaluate different attentive listening systems. These systems differ in the level of control provided to the operator and system. We define these systems as fully operator (\textbf{HUMAN}), fully autonomous (\textbf{AUTO}) and semi-autonomous (\textbf{SEMI}). This work determines how the \textbf{SEMI} system fares in relation to the former two systems as a continuum of the level of human involvement. 

In the \textbf{HUMAN} experiment, 20 subjects underwent an attentive listening session for eight minutes with a remote human operator who was speaking through the agent. Subjects were given some time before the experiment to consider what they were going to talk about. They could choose any topic they wished, such as their recent holiday or an interesting thing that happened to them. The remote operator was instructed to perform the role of an attentive listener while also controlling the facial expressions of the agent. The interface provided is the same as in Figure \ref{op_screen}, except that the operator could not turn off the microphone - they would talk continuously to the user who would not interact with any part of the autonomous system. Users were informed before the experiment that they would be conversing with a remote operator. This operator was different than the operator used for the \textbf{SEMI} experiment.

For the \textbf{AUTO} experiment, a separate 20 subjects were asked to participate in attentive listening using the same procedure as \textbf{HUMAN}, but with the fully autonomous system described in Section \ref{att_list}. We use this experiment as a baseline for improvement in future iterations of the system.

Another 20 subjects participated in the final experiment condition with the \textbf{SEMI} system, where a remote operator can take over from the autonomous agent, as described earlier in this work. Subjects in this experiment were also previously informed that an operator would be listening in and have the ability to take over the interaction, though they were not informed about the conditions when this would take place. The voice of the operator was modified to make a voice closer to the TTS voice of the agent, although it was not an exact match and the change in voice was obvious to the subject.

If one of the conditions of takeover in the \textbf{SEMI} system was reached, the interface displayed a message to the operator asking them to take control of the agent. The operator was informed to take control and say a short utterance to keep the conversation on track. The operator was instructed not to converse with the subject for a long period of time over multiple turns. The operator was bound by the conditions of interruption and only allowed to take over when the system prompted them to do so.

All subjects were university students. They were compensated for their participation in the experiment and signed consent forms to record their data for analysis.

\section{Results}
For all experiments the subjects evaluated 19 items using a 7-point Likert scale with the extrema items as ``complete disagreement'' (1) and ``complete agreement'' (7). We used Ward's clustering over the results of all sessions to group these items into broader measures as displayed in Table \ref{questionnaire}.

\begin{table}
\caption{Grouping of questionnaire items into measures}
\label{questionnaire} 
\begin{tabular}{ll}
\hline\noalign{\smallskip}
Measures & Item  \\
\noalign{\smallskip}\svhline\noalign{\smallskip}
Naturalness & The agent's responses were human-like \\
						& The words the agent used were natural \\
						& The agent's responses could stimulate my own talk \\
						& The agent understood my talk \\
\noalign{\smallskip}\hline\noalign{\smallskip}
User satisfaction & The agent was easy to talk to \\
& I want to talk with the agent again\\
& The conversation was smooth\\
& I was satisfied with the conversation\\
\noalign{\smallskip}\hline\noalign{\smallskip}
Utterance Timing & The agent responded at an appropriate frequency\\
& The agent responses were well timed\\
& The agent had good pauses in the conversation\\
\noalign{\smallskip}\hline\noalign{\smallskip}
Empathetic listening & The agent displayed empathy towards me\\
& The agent took the conversation seriously\\
& The agent was listening intently\\
& The agent was listening actively\\
& The agent was accommodating\\
\noalign{\smallskip}\hline\noalign{\smallskip}
Interest & The agent showed interest in the conversation\\
& The agent responded with special care\\
\noalign{\smallskip}\hline\noalign{\smallskip}
Other (omitted from analysis) & The agent was fully autonomous\\
\noalign{\smallskip}\hline\noalign{\smallskip}
\end{tabular}
\end{table}





\begin{figure}[t]
\centering
\includegraphics[width=0.95\columnwidth]{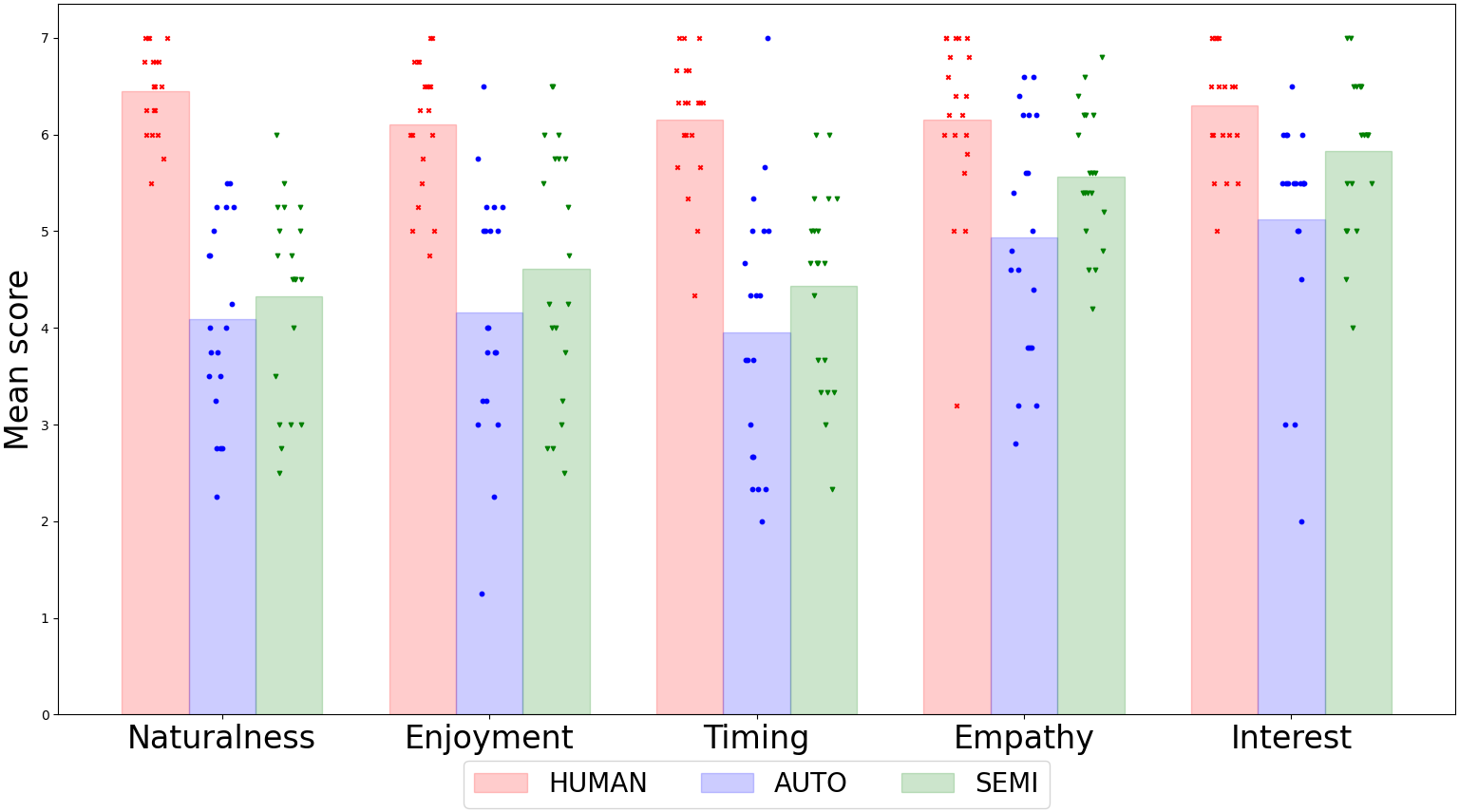}
\caption{Average scores for each measure across the three conditions. Sample points are given for each of the 20 subjects per condition.}
\label{bar} 
\end{figure}

In this work we use the mean of the items in the measures in Table \ref{questionnaire} as the basis for comparison. \textbf{HUMAN}, \textbf{AUTO} and \textbf{SEMI} conditions were first compared using ANOVA and it was found that there were differences between one or more groups. We then conducted post-hoc t-tests between each individual group. Results are shown in Figure \ref{bar}.

It was found that the \textbf{HUMAN} condition was rated higher than the other two conditions across all measures, which is natural considering it is the upper baseline. Across all measures, the \textbf{SEMI} system outperforms the \textbf{AUTO} system, although to varying degrees. For naturalness, enjoyment, and timing the two systems do not show statistically significant differences although the average score for \textbf{SEMI} is higher. Significant differences are seen between \textbf{SEMI} and \textbf{AUTO} in the empathy and interest measures ($p\leq0.05$).

One interesting aspect that can be seen from the individual data points in Figure \ref{bar} is that the \textbf{SEMI} system appears to improve subject evaluations on the lower end of the scale. That is, while subjects with a comparatively high rating are fairly equal between \textbf{SEMI} and \textbf{AUTO}, subjects who rated the \textbf{AUTO} system lower were more harsh, particularly for the empathy measure.

In terms of operator takeovers, the median number per session was 4.5, with the range between 0 and 14 within the eight-minute sessions. From the 20 subjects, 4 did not experience any takeover from the operator. The average time the operator spent speaking to the user during a session was 18.5 seconds, with an average of three seconds speaking time per takeover, not including silence time before and after the operator switched. In some cases, the operator had to wait for an appropriate moment to interject before speaking.

We also analyzed the correlation between the frequency of takeovers and each of the measures. Results are shown in Table \ref{correl}.  

\begin{table}
\centering
\caption{Correlation coefficient between measures and number of takeovers}
\label{correl} 
\begin{tabular}{ll}
\hline\noalign{\smallskip}
Measure & $R^{2}$ value  \\
\noalign{\smallskip}\svhline\noalign{\smallskip}
Naturalness & -0.08 \\
Enjoyment & -0.32 \\
Timing & -0.11 \\
Empathy & -0.10 \\
Interest & -0.12 \\
\end{tabular}
\end{table}

\begin{figure}[h]
\centering
\includegraphics[width=0.95\columnwidth]{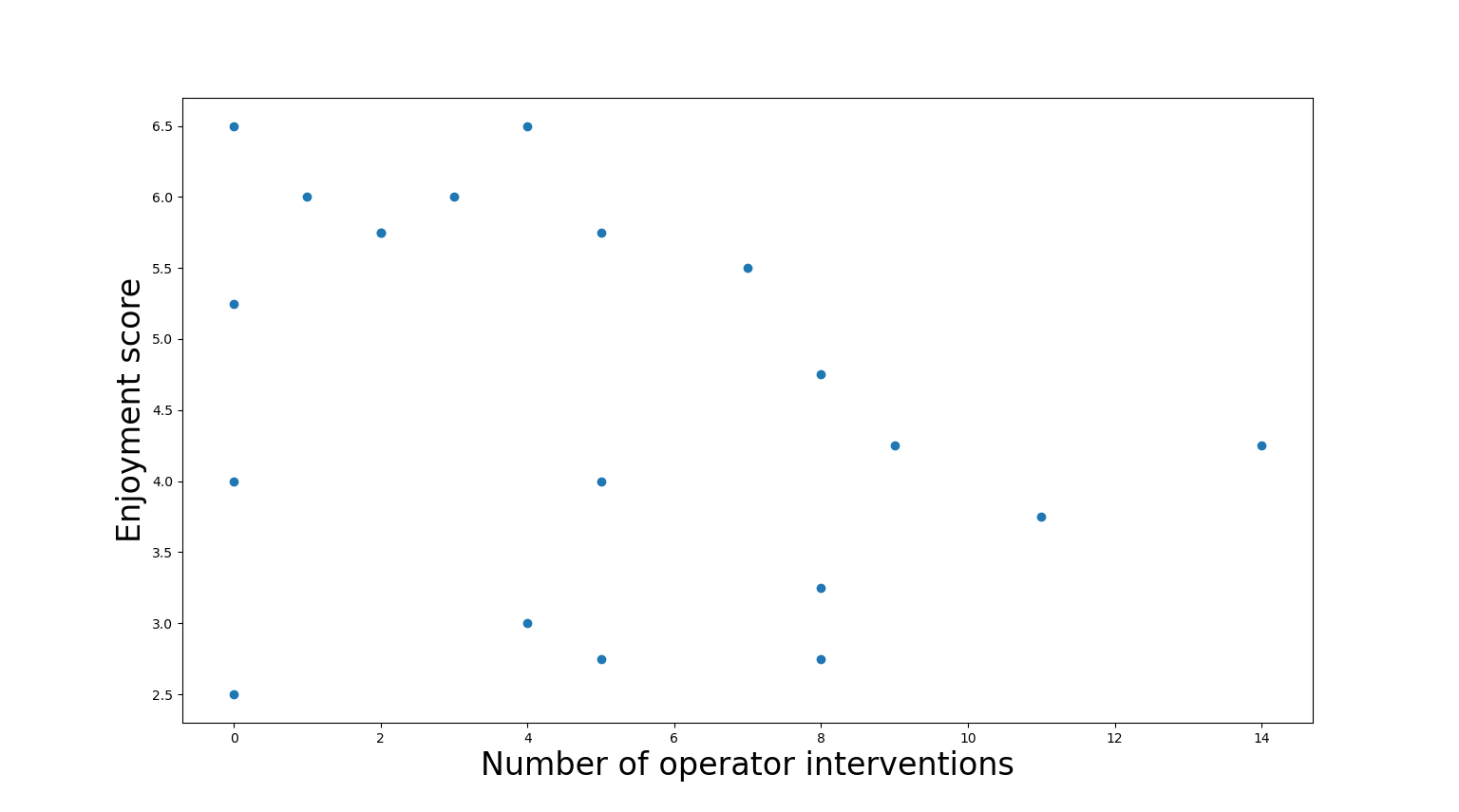}
\caption{Average scores for each measure across the three conditions. Sample points are given for each of the 20 subjects per condition.}
\label{enj_correl} 
\end{figure}

We find that all of the measures have negative correlation with the number of operator takeovers, however these values are small in absolute value which indicates this correlation is minimal. For enjoyment, the correlation of -0.32 was the most significant, however we find from the scatter plot in Fig. \ref{enj_correl} that there is no obvious pattern.

From this analysis we can conclude that the perception of the agent is not greatly influenced by the number of times the operator takes control of the system. Furthermore, we directly asked subjects if they felt the frequency of operator takeovers was appropriate. From the 16 subjects who experienced takeovers, 11 said they felt the frequency of operator takeovers was appropriate, 2 said there were not enough takeovers, while 3 said there were too many takeovers.

Subjects also answered additional questions related to the timing and effectiveness of the operator takeovers on a 7-point Likert scale. For timing, 10 subjects agreed that the timing of the takeovers was appropriate, 2 were neutral, while 4 disagreed. For the question of whether the takeovers were effective in improving the conversation, 12 subjects agreed, 1 subject was neutral, and 3 disagreed. In general, takeovers improved attentive listening. 

\section{Discussion}
From our experiments we conclude that a semi-autonomous system can improve the agent by reducing the number of users who have a comparatively negative impression of the system. Attentive listening is not an everyday dialogue situation and so many users find it difficult to engage in an such a scenario. We have shown that the addition of a human to ``rescue'' a conversation can improve the overall dialogue. This assistance is more useful for users who are unable to converse with an attentive listening system ``correctly''. For users who can do this well, the performance of the autonomous dialogue system itself is likely to be more influential on their perception, although we did not confirm this here.

Although the \textbf{SEMI} system outperformed the \textbf{AUTO} system, we only found statistically significant differences over certain measures. From the perspective of attentive listening, the improved performance of empathy and interest show that a human can display these traits better than our dialogue system could. However, the differences across other measures were not so marked. In particular, the similarity in scores across naturalness and timing measures could be explained by the novelty of the interaction (a human taking over control of a conversation at certain times). This type of interaction was unfamiliar for subjects and so could be seen as somewhat unnatural. Similarly, the timing of operator responses when deciding to take over and beginning to speak may feel somewhat jarring for the subjects. In particular, the modified voice of the operator was not the same as the actual TTS of the autonomous agent.

Importantly, we have also shown that this improvement is not merely a function of the frequency of intervention. It means that the decision to take over should be made judiciously. Our pilot experiments also showed that a human having free reign over intervention is not desirable. Therefore, our proposal that the system should make the decision for the human to intervene seems to be somewhat justified. The main issue is then how the system can accurately recognize the conditions of this intervention. 

The conditions in which the system recognizes that it should hand over control to a human were formulated by observing and analyzing previous interactions. Our aim is to choose conditions which are explicit and understandable for the human operator so the decision to take over is not a ``black box''. However, it is likely that there are many implicit signals which we did not capture that indicate that the user is losing engagement. These may include things like a disinterested tone of voice or gaze behavior, or even the specific content of the talk. We acknowledge that there is much scope to improve the detection of low engagement which may be served by more powerful machine learning algorithms. On the other hand, we feel that making the decision to take over explainable to the operator is a necessary feature of the semi-autonomous system. 

The nature of our attentive listening system allowed us to estimate low engagement since the patterns of responses are known by the system. The system knows the diversity of responses which it generates and so can understand when there is an imbalance in the way that it communicates to the user, for example with too many uninteresting  formulaic responses. Other types of dialogue systems may not function in such a manner, although may be useful for a system to understand the diversity of its responses so that it can avoid repeating the same patterns during the conversation.

One limitation of this work was that the operator of the \textbf{SEMI} system was the same for every experiment and was familiar with the goal of the experiment. Although this may introduce some bias, the operator was still bound by the system, in that they could not take over unless the system had detected a heuristic for take over. The quality of the takeover would also logically depend on the skill of the operator. We wish to use skilled experts in conversation as operators to examine this effect. 

Future work in this area is to now apply these results to our autonomous system. Now that the system can detect points of low interest through heuristics, we can focus on such segments and generate dialogue which deals with them specifically. For example, a user whose turns are too short should be asked a more complex question, which may require assistance from large language models rather than the template responses which have been used in our baseline autonomous system.

Another extension of this work is to apply the semi-autonomous system to not only one-to-one conversation, but to allow one operator to handle multiple conversations in parallel. We have recently began to conduct experiments of this nature, with the intention of showing that one operator can effectively monitor many users simultaneously without suffering from a large cognitive load.

\section{Conclusion}
In this work we developed a semi-autonomous attentive listening system where a remote operator could take over control from an autonomous agent. The system prompts the user to take over when it detects low engagement from the user during the conversation. The operator then takes over and can speak directly to the user through the agent. We compared this system to a fully autonomous and fully tele-operated system across several measures. It was found that the semi-autonomous system could improve empathy and interest of the fully autonomous system. We also find that operator takeovers generally resulted in a positive impression on users and this was independent of the frequency of the takeovers. Future work includes integrating the conditions into an upgraded autonomous system for iterative improvement.

\begin{acknowledgement}
This work was supported by JST Moonshot R\&D Goal 1 Avatar Symbiotic Society Project (JPMJPS2011).
\end{acknowledgement}

\bibliographystyle{spmpsci}
\bibliography{IWSDS2024_bib}

\end{document}